\documentclass[letterpaper, 10 pt, conference]{ieeeconf}  

\IEEEoverridecommandlockouts                              

\usepackage{times}
\usepackage{multicol}
\usepackage{latexsym}
\usepackage{color}
\usepackage{amsmath}
\usepackage{cite}     
\usepackage{listings}
\usepackage{mathrsfs}
\usepackage{amssymb}
\usepackage{pstool}
\usepackage{psfrag}
\usepackage[ruled]{algorithm2e}
\usepackage{hyperref}
\usepackage[table]{xcolor}
\usepackage{graphicx}
\usepackage{amsthm}
\usepackage{color,soul}
\usepackage{xcolor}
\usepackage{lineno,hyperref}
\usepackage{epsfig}
\usepackage{subfigure}
\usepackage{amssymb}
\usepackage{siunitx}
\usepackage{latexsym}
\usepackage{amsmath}
\usepackage{amsfonts}
\usepackage{blindtext}
\usepackage{scrextend}
\usepackage{pbox}
\usepackage{xcolor}
\usepackage{svg}
\usepackage[most]{tcolorbox}
\usepackage{amsthm}
\usepackage{float}

\theoremstyle{definition}
\newtheorem{definition}{Definition}

\theoremstyle{remark}
\newtheorem{remark}{Remark}

\theoremstyle{assumption}
\newtheorem{assumption}{Assumption}

\title{\LARGE \bf
Embodied Hazard Mitigation using Vision-Language Models for Autonomous Mobile Robots}

\author{Oluwadamilola Sotomi$^{1}$, Devika Kodi$^{1}$, Kiruthiga Chandra Shekar$^{1,2}$, and Aliasghar Arab$^{*1,2}$
\thanks{This work was supported by the Department of Mechanical and Aerospace Engineering, New York University Tandon School of Engineering, and a Nokia Bell Labs Research Award.}%
\thanks{$^{1}$Oluwadamilola Sotomi, Devika Kodi, Kiruthiga Chandra Shekar, and Aliasghar Arab are with the Department of Mechanical and Aerospace Engineering, Tandon School of Engineering, New York University, 6 MetroTech Center, Brooklyn, NY 11201, USA.  
{\tt\small ots2014@nyu.edu, dak9250@nyu.edu, kcs433@nyu.edu, aliasghar.arab@nyu.edu}}%
\thanks{$^{2}$Kiruthiga Chandra Shekar and Aliasghar Arab are with GenAuto.ai by General Autonomy Inc.  
{\tt\small kshekar@genauto.ai, mojarab@genauto.ai}}%
}
\begin{document}
\maketitle

\begin{abstract}
Autonomous robots operating in dynamic environments should identify and report anomalies. Embodying proactive mitigation improves safety and operational continuity. This paper presents a multimodal anomaly detection and mitigation system that integrates vision-language models and large language models to identify and report hazardous situations and conflicts in real-time. The proposed system enables robots to perceive, interpret, report, and if possible respond to urban and environmental anomalies through proactive detection mechanisms and automated mitigation actions. A key contribution in this paper is the integration of Hazardous and Conflict states into the robot's decision-making framework, where each anomaly type can trigger specific mitigation strategies. User studies (n = 30) demonstrated the effectiveness of the system in anomaly detection with 91.2\% prediction accuracy and relatively low latency response times using edge-ai architecture.
\end{abstract}

\section{INTRODUCTION}
\color{black}{
As Autonomous Mobile Robots (AMRs) are increasingly deployed in dynamic and unpredictable environments, the ability to proactively detect and respond to anomalies is critical for ensuring safety and operational continuity~\cite{baraglia2017efficient}. Traditional AMRs operate with limited situational awareness, often reacting to problems only after they occur, which can lead to safety incidents, operational disruptions, and reduced system reliability~\cite{shekar2024explainable, lee2004trust}. A key requirement in autonomous robotics is therefore proactive anomaly detection—robots must not only navigate safely but also anticipate and proactively respond to hazards for mitigation before they escalate into critical situations~\cite{sanneman2022situation}.
Past work has explored reactive safety systems and obstacle avoidance~\cite{laban2023opening}, but many approaches remain restricted to predefined scenarios, lacking adaptability to new anomalies or changing environments. The early models relied on rule-based or threshold detection methods~\cite{sobrin2024enhancing}, which did not capture contextual information or provide meaningful responses. More recent machine learning approaches combine visual perception and semantic reasoning to improve anomaly detection~\cite{abbas2024talkwithmachines, das2021explainable, zhang2023anomaly}.

\begin{figure}[ht!]
\centering
\includegraphics[width=.9\linewidth]{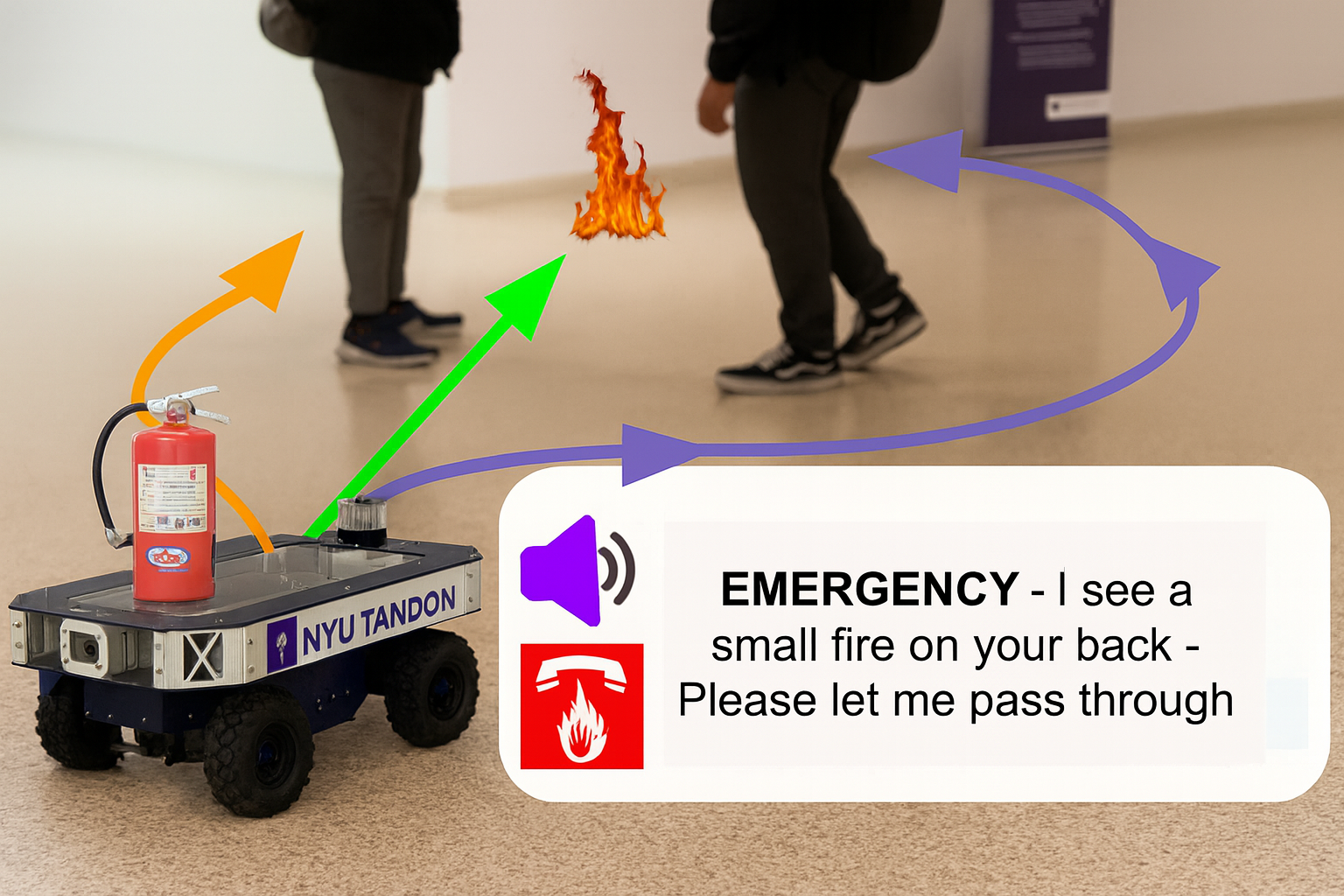}
\caption{The AMR detects abnormalities during normal operation and proactively responds through real-time reporting and mitigation planning to ensure safety.}
\label{fig:social}
\end{figure}

Recent advances in large language models (LLMs) and vision language models (VLMs) open new opportunities for intelligent anomaly detection~\cite{gavriilidis2023surrogate, chen2024llm}. These models process multimodal inputs, understand the context, and generate human-like reasoning about hazards and conflicts~\cite{finkelstein2022explainable}. Previous studies emphasize aligning robot behavior with human safety expectations to support trust in real-world deployments~\cite{singh2024towards, cruz2022evaluating, liu2023safety}. In particular, VLMs improve AMR perception by integrating visual and textual understanding: Grad-CAM~\cite{selvaraju2020grad} highlights salient regions, VLFM~\cite{yokoyama2024vlfm} allows semantic mapping, and BLIP~\cite{li2022blip} grounds image–text pairs for hazard identification. These capabilities have been extended to explainable AI safety frameworks that link anomaly detection with autonomous decision making~\cite{hamilton2020autonomous, wang2024vlm, patel2024mitigation}.
To address current limitations, we propose a proactive anomaly detection and mitigation system that generates real-time hazard assessments and automated responses for AMR operations. Using VLFM and LLM, our approach integrates camera-based perception, semantic analysis, and language-based reasoning for contextual anomaly detection. A key innovation is the explicit modeling of two distinct anomaly states, Hazardous and Conflict, each linked to targeted mitigation strategies. This work extends our previous research on explainable safety systems~\cite{shekar2024explainable, doma2024llm} with expanded anomaly detection capabilities and a real-time ROS2 implementation using AI models on the edge. Our results demonstrate improved safety, faster response, and more transparent decision-making, underscoring the potential of VLFMs and LLMs to enhance operational reliability of AMRs in dynamic environments.


\section{PROBLEM FORMULATION}
Autonomous mobile robotic systems operating in dynamic environments must proactively detect and respond to anomalies to ensure safety and operational continuity. We define the proactive anomaly detection and mitigation task as a tuple of perception information from the data in the robot world.
\begin{equation}
\label{eq:tuple}
\mathcal{T}_{\text{robot}} = \left( \mathcal{S}, \mathcal{G}, \mathcal{P}, \mathcal{A}, \mathcal{M}, \mathcal{H}, \mathcal{L}, \mathcal{R}, \varepsilon \right)
\end{equation}

\noindent where, \( \mathcal{S} = (\mathbf{q}, \mathbf{v}, \mathbf{q}_{\text{env}}) \) is the state of the robot, with \( \mathbf{q} \in \mathbb{R}^n \) as the position and orientation of the robot, \( \mathbf{v} \in \mathbb{R}^n \) as the velocity of the robot, \( \mathbf{q}_{\text{env}}^j \in \mathbb{R}^n \) represents the relative state of an observed environmental object. These objects might represent a hazardous situation from the robot's point of view and can be defined. \( \mathcal{G} \equiv \mathbf{q}_g \in \mathbb{R}^n \) is the goal in the robot workspace. \( \mathcal{P} = \pi: [0, T] \rightarrow \mathbb{R}^n \) is the planned trajectory that maps time to robot location and velocities, so that the robot safely transitions from the initial state \( \mathbf{q}_0 \) to \( \mathbf{q}_g \) while avoiding detected anomalies and maintaining operational safety. \( \mathcal{A} = \{ a_t \mid t \in [0, T] \} \) is the set of anomalies detected during execution, where each \( a_t \) includes the type of anomaly, severity assessment, the state of the robot, the relative state of the objects, and contextual information through multimodal analysis.

\noindent \( \mathcal{M} = \{ m_t \mid t \in [0, T] \} \) is the set of mitigation actions triggered in response to detected anomalies, where each \( m_t \) represents automated responses such as path replanning, emergency stops, or alert notifications. \( \mathcal{H} \subset \mathcal{W} \) represents the subset of known hazardous states in which the robot is previously exposed to risk of loss, with \( \mathcal{W} \) being the set of all possible world states. \( \mathcal{L} \subset \mathcal{W} \) represents the subset of loss states in which accidents or undesirable results occur. \( \mathcal{R} = \{ R_i(\mathbf{S}_h) \mid i \in [1, n] \} \) is the set of risk functions that map hazardous states to loss probabilities, where each \( R_i(\mathbf{S}_h) \in [0, 1] \) represents the probability of transitioning from a hazardous state to a specific loss class. \( \varepsilon \in [0, 1] \) is the anomaly detection effectiveness score that reflects the system's ability to identify and respond to threats, measured through detection accuracy and response time metrics.

The set of safety constraints, anomaly detection requirements, and mitigation protocols can be formalized as a set of operational constraints \( \Omega_{safety} \), which must be satisfied at all times.

\begin{equation}
\Omega_{\text{safety}} = \bigcap_{i \in M} \Omega_i
\end{equation}

\noindent where \( \Omega_i \) represents the constraints imposed by the safety requirement \( i \) from the set of governing rules \( M \). For this purpose, which can be modeled in different categories as suggested in~\cite{arab2021safe}.

\subsection{Hazard as Risk for Loss}
Any state that has a risk or probability of transitioning to a loss state is a hazardous state. In other words, an autonomous robot at any given time of $t \in \mathbb{R}>0$ can be represented by the state $\mathbf{S}$ that belongs to the general world $\mathbf{S} \in \mathcal{W}$, where $\mathcal{W}$ represents the set of all possible states. Even though all feasible world states are not known, this formulates the problem by separating known and unknown subsets of the world. Let us also assume that there exists a subset of \textit{Hazardous} states, named $\mathcal{H} \subset \mathcal{W}$ where the robot is in a state exposed to risk of loss. Similarly, in practice, all these states are not identified and an unknown subset of hazardous states might exist. Hence, this set is dynamic and, if any loss occurs, the states that ended up in a loss event should be added to $\mathcal{H}$, the hazardous subset.

\begin{remark}[\textbf{Robot World}] Representing the robot's world $\mathcal{T}_{\text{robot}}$ in the universal state space $\mathcal{W}$ and state $\mathbf{S}$ that is dependent on the autonomy architecture, capabilities and limitations of the robot's perception technology and the context at the scenario. However, for the purposes of formulating this problem, the world as perceived from the autonomous robot's point of view consists of the spatio-temporal information of the robot and relative items in its surroundings.
\end{remark}

\begin{definition}[\textbf{Hazardous State}]
A state $\mathbf{S}$ is identified as hazardous state $\mathbf{S}_h$ if $\mathbf{S}_h \in \mathcal{H}$ exists, which means that there is a potential source of harm with the possibility of leading to a loss. The source might be originating from the malfunctioning of a system component or exposure to world states.
\end{definition}

\begin{definition}[\textbf{Risk of loss}]
The risk of loss is the likelihood that a hazardous state ends up in a loss state in a limited time. A probabilistic function $R_i(\mathbf{S}_h) \in [0,\:1]$ can represent the likelihood of that specific incident. This risk assessment can also be utilized for explainability purposes, allowing the system to communicate the probability of different loss scenarios to users and operators, thus improving transparency in decision-making processes.
\end{definition}

\section{METHODOLOGY}
The objective is to calculate a safe, feasible and anomaly-aware path \( P \), while maximizing \( \varepsilon \) through novel anomaly detection and mitigation modules, to improve safety and operational reliability during robot navigation in dynamic environments. Our approach consists of four parts.
\begin{itemize}
    \item \textbf{1)} Development of a multimodal anomaly detection system using VLM and LLM.
    \item \textbf{2)} Deployment in an AMRs for real-time validation.
    \item \textbf{3)} Integration with an autonomous navigation stack.
    \item \textbf{4)} Integration of proactive mitigation strategies for hazardous states.
\end{itemize}

\begin{assumption}
The effectiveness of the anomaly detection module is quantified by a scalar \emph{anomaly detection factor} \( \varepsilon \in [0, 1] \), which reflects how well the robot can identify and respond to environmental threats. The value of \( \varepsilon \) is determined through performance metrics that include detection accuracy, false positive rates, and response time measurements.

\begin{equation}
\varepsilon =
\begin{cases}
0, & \text{if anomaly detection is inactive}, \\
\hat{\varepsilon} \in (0, 1], & \text{if anomaly detection is active}.
\end{cases}
\end{equation}
\noindent where, \( \hat{\varepsilon} \) is a normalized score derived from detection performance metrics and response time evaluations.
\end{assumption}

\begin{algorithm}[h!]
\caption{Anomaly Detection and Mitigation Module via VLM, Heatmap and LLM}
\label{alg:AnomalyDetection}
\nl Initialize LLM Node and Anomaly Detection Module\;
\nl Subscribe to topics \texttt{'camera/image'}, \texttt{'blip/caption'}, and \texttt{'heatmap/summary'}\;
\nl Set anomaly detection factor \( \varepsilon \leftarrow 0 \)\;
\nl Set AnomalyDetectionModuleEnabled flag\;

\While{robot is navigating}{
    \nl Receive image from camera stream\;
    \nl Detect potential anomalies using VLM Node\;
    \nl Generate visual saliency map using Heatmap Node\;
    \If{anomaly is detected}{
        \If{AnomalyDetectionModuleEnabled}{
            \nl Classify anomaly as Hazardous or Conflict using LLM Node\;
            \nl Trigger appropriate mitigation action based on classification\;
            \nl Generate natural language description of detected anomaly\;
            \nl Overlay and display heatmap with anomaly classification\;
            \nl Save image, heatmap, and anomaly report with timestamp\;
            \nl Update anomaly detection factor \( \varepsilon \leftarrow \varepsilon + \Delta\varepsilon \)\;}
        \nl Update navigation path to avoid detected anomaly\;}
    \nl Execute current navigation step\;}
\nl Analyze anomaly detection performance metrics (e.g., detection accuracy, response time)\;
\nl Correlate performance with anomaly detection factor \( \varepsilon \)\;
\end{algorithm}

\subsection{Anomaly Detection Model}
The robot is equipped with a modular anomaly detection model implemented as four ROS2 nodes, 1) Camera node, 2) BLIP node, 3) Heatmap node and 4) LLM node which will be explained in the experimental section. Each node is responsible for a distinct function. These nodes communicate through ROS topics, enabling scalable and seamless integration with existing navigation systems. This node presents information in a structured format for the classification and mitigation of anomalies, enhancing safety in dynamic environments. The camera captures a single image on request, the LLM node must be initialized first, followed by the Heatmap and BLIP nodes.

\subsubsection{Anomaly Detection Model Formulation}
To formally define our anomaly detection model, let \( X \) represent the raw image input captured by the robot camera. The anomaly detection function \( A \) maps the visual input, the heatmap analysis, and the language model output to a structured anomaly classification by:

\begin{equation}
A: (X, H, L) \rightarrow \mathcal{C}
\end{equation}

\noindent where, \( X \in \mathbb{R}^{m \times n \times 3} \) is the image captured at resolution \( m \times n \),  \( H = g(X) \) is the heatmap function that highlights the salient regions,  \( L = f(X, H) \) represents the captioning output of the language model and \( \mathcal{C} \) is the final anomaly classification produced. The heatmap generation function \( g(X) \) is given by Grad-CAM activation \( A_c \) as

\begin{equation}
H_{i,j} = ReLU\left( \sum_k \alpha_k A_c^{i,j} \right)
\end{equation}

\noindent where, \( \alpha_k \) is the weight for the feature map \( k \), \( A_c^{i,j} \) represents the activation at the spatial location \( (i,j) \), and \( ReLU(\cdot) \) ensures positive activation contributions. The final anomaly classification \( \mathcal{C} \) is derived using

\begin{equation}
\mathcal{C} = LLM(\psi (H, X))
\end{equation}

\noindent where, \( \psi(H, X) \) is the feature representation that combines the heatmap and the image context and \( LLM(\cdot) \) is a large language model (e.g. Generative Pre-trained Transformer 3.5 Turbo) trained for anomaly classification obtained in the LLM Anomaly Detection Prompt box. The anomaly classification generated by the LLM, which depends on the environmental context and perception input and the variables related to the robot interface, captured as uncertainty $\mathcal{U}$, which reflects the confidence of detection, the accuracy of the classification and the reliability of the response.

\begin{equation}
\varepsilon = f\left( \mathcal{C}, \mathcal{U} \right).
\label{eq:anomaly_detection_factor}
\end{equation}
Performance metrics will allow us to determine $\varepsilon$ more precisely.

\subsubsection{Latency Optimization for Real-Time Anomaly Detection}
Latency is critical in real-time safety systems. The total anomaly detection time \( T_{\text{total}} \) is defined as:

\begin{equation}
T_{\text{total}} = T_{\text{camera}} + T_{\text{BLIP}} + T_{\text{heatmap}} + T_{\text{LLM}}
\end{equation}

\noindent where, \( T_{\text{camera}} \) is the image acquisition time, \( T_{\text{BLIP}} \) is the processing time in the vision language, \( T_{\text{heatmap}} \) is the heatmap generation time, and \( T_{\text{LLM}} \) is the time required for the large language model to generate an anomaly classification. Since LLM processing is performed remotely, LLM request latency \( T_{\text{LLM}} \) can be modeled as

\begin{equation}
T_{\text{LLM}} = T_{\text{network}} + T_{\text{processing}}
\end{equation}

\noindent where, \( T_{\text{network}} \) represents the latency of network transmission and \( T_{\text{processing}} \) is the cloud-based inference time. To minimize \( T_{\text{total}} \), one can formulate the optimization problem as

\begin{equation}
\min_{\lambda} \sum_{i} T_i, \quad \text{s.t.} \quad T_{\text{total}} \leq T_{\text{max}}
\end{equation}

\noindent where, \( \lambda \) represents hyperparameters tuning latency trade-offs and \( T_{\text{max}} \) is the maximum allowable latency for real-time safety operation.

Empirical analysis showed that latency is inversely correlated with compute power \( C \):

\begin{equation}
T_{\text{total}} \propto \frac{1}{C}
\end{equation}

\noindent where increasing computing power reduces processing time.

\begin{figure*}[ht!]
    \centering
    \includegraphics[width=0.9\linewidth]{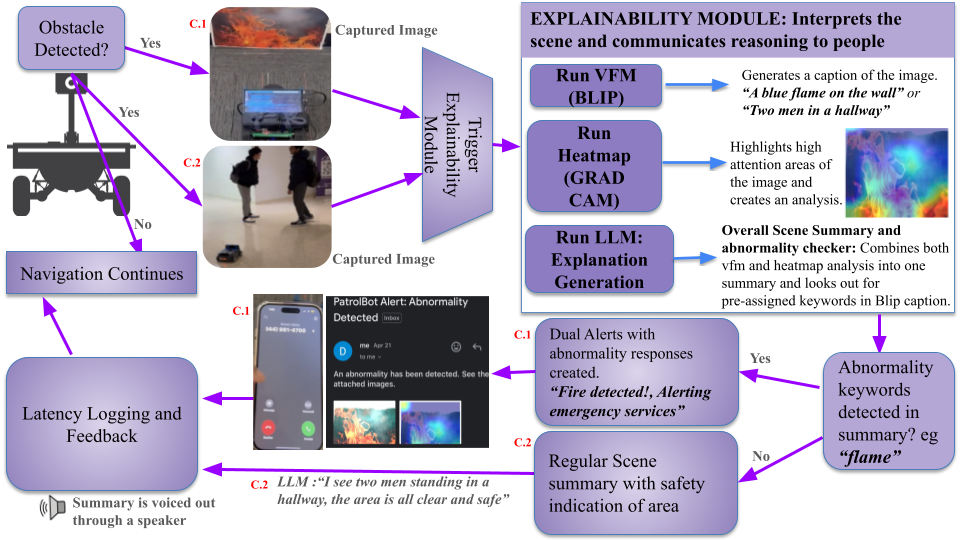}
    \caption{Workflow of anomaly detection and explanation and emergency call.}
    \label{fig:placeholder}
\end{figure*}

\section{EXPERIMENTS}
To validate the effectiveness of our anomaly detection and mitigation module, we conducted structured experiments using a mobile robot running ROS 1 Noetic on a Raspberry Pi 4B with a built-in camera. The robot ROS 1 autonomy stack can communicate with the ROS 2 abnormally module. However, we tested manual and autonomous navigation scenarios with various types of anomaly to evaluate detection accuracy and response effectiveness.

\subsection{Experimental Setup}
The experiments were carried out in an academic indoor environment that included a makerspace lab and a hallway (2.5m × 14m) to simulate real-world deployment conditions with controlled anomaly scenarios. The anomaly detection module, originally built in ROS 2 Humble, was adapted to ROS 2 Foxy and deployed separately for compatibility with the MYAGV robot. The images were captured every 5 seconds and processed through the Camera, BLIP, Heatmap, and LLM nodes, with real-time anomaly classifications delivered via heatmap overlays and automated mitigation actions.

\begin{tcolorbox}[colback=gray!10!white, colframe=black, title=LLM Anomaly Detection Prompt, sharp corners=south, boxrule=0.5mm]
\textit{
"You are a mobile robot that monitors your environment for potential hazards and conflicts. The image caption is: '\{caption\}'. The heatmap analysis shows: '\{heatmap\_summary\}'. Analyze this information and classify any detected anomalies. If you detect a hazard (e.g., dangerous objects, environmental threats, safety violations), respond with 'HAZARDOUS: [brief description]' and 'REPORT'. If you detect a conflict (e.g., navigation conflicts, operational disruptions, rule violations), respond with 'CONFLICT: [brief description]' and 'AVOID'. If no anomalies are detected, respond with 'CLEAR: [brief description of normal conditions]' and 'RESUME'."}
\end{tcolorbox}

\subsection{Navigation and Testing Conditions}
The experiments were carried out under four scenarios:
\begin{itemize}
    \item \textbf{Manual Navigation:} As Test 1 - With and without anomaly detection.
    \item \textbf{Autonomous Navigation:} As Test 2 - With and without anomaly detection.
\end{itemize}
\noindent For each test, we recorded navigation metrics and anomaly detection output, allowing us to isolate the impact of proactive anomaly detection as represented in Table~\ref{tab:navigation_metrics}.

\begin{table}[h!]
\centering
\begin{tabular}{|l|c|c|}
\hline
\textbf{Metric} & \textbf{WoAD} & \textbf{WAD} \\
\hline
\multicolumn{3}{|c|}{\textbf{Test 1: Manual Navigation}} \\
\hline
Total Trajectory (m) & 5.76 & 5.76 \\
Total Time (s) & 23.5 & 22.1 \\
Anomalies Detected & -- & 2 \\
Sudden Stops & 19 & 15 \\
\hline
\multicolumn{3}{|c|}{\textbf{Test 2: Autonomous Navigation}} \\
\hline
Total Trajectory (m) & 5.83 & 5.78 \\
Total Time (s) & 25.3 & 22.6 \\
Anomalies Detected & -- & 3 \\
Sudden Stops & 21 & 18 \\
\hline
\end{tabular}
\caption{Navigation performance comparison: WoAD = Without Anomaly Detection, WAD = With Anomaly Detection. Results are averaged over four 14-meter delivery runs. Metrics include trajectory length, time, detected anomalies, and sudden stops.}
\label{tab:navigation_metrics}
\end{table}


\subsection{Anomaly Detection and Mitigation Evaluation}
To extend the framework toward comprehensive anomaly detection, the LLM node was augmented with a response mapping layer. This module scanned captions for predefined anomaly keywords (e.g. firearms, people fighting, obstruction, hazard) and assigned each to a specific mitigation action, such as issuing a verbal warning, triggering a simulated phone call or email, or activating a siren shown in Table~\ref{tab:anomalies_mitigation}. When no anomaly keywords were found, the LLM defaulted to a normal environment assessment, ensuring continuous monitoring during routine operations.

For testing, we simulated anomalies by placing large printed images (bags, hazard signs, blockages) along the robot's route. These images dominated the robot's field of view and reliably triggered the anomaly detection pipeline. Multiple checkpoints ensured that the consistency of detection and response could be validated across various scenarios.

\begin{table}[h!]
\centering
\caption{Anomalies and Mitigations}
\label{tab:anomalies_mitigation}
\begin{tabular}{|c|c|c|}
\hline
\textbf{Hazard} & \textbf{Level} & \textbf{Mitigation} \\ \hline
Firearm & High & Siren, notify \\ \hline
Fight & High & Warn, alert \\ \hline
Obstruction & Medium & Replan path \\ \hline
Spill & Medium & Report, avoid \\ \hline
None & Low & Resume ops \\ \hline
\end{tabular}
\end{table}

\subsection{Latency and Prediction Accuracy Evaluation}

We measured the performance of the patrol extension in terms of responsiveness and precision. Latency, defined as the time from image capture to explanation or alert, ranged from 2.5 to 25.2 s, with 84\% of the responses occurring within 14 s. The prediction accuracy was calculated by comparing the generated responses with ground-truth anomaly labels: 114 correct detections of 125 trials yielded a precision of 91.2\%. Finally, we validated the auxiliary response chains. Simulated phone calls, email alerts, and siren activations were all successfully triggered once anomalies were identified, confirming that the patrol system could extend beyond explanation into actionable responses.

\section{ANALYSIS}
We analyze AMR performance metrics along with \( \varepsilon \) to assess how anomaly detection influenced navigation behavior and safety outcomes. This included latency, detection accuracy, and confusion matrix evaluations that compared the system output with the ground-truth anomaly labels. Table~\ref{tab:survey_results} summarizes the responses to test 2, showing a significant improvement in safety, trust and operational reliability when anomaly detection was enabled. We computed the overall preference score using the following.

\begin{equation}
\text{PS} = \frac{U + 0.5N}{T} \times 100,
\label{eq:preference_score}
\end{equation}

\noindent where, \( U = 22 \) (users who prefer anomaly detection), \( N = 6 \) (neutral responses), and \( T = 30 \) (total participants), resulting in a PS of $83.3\%$. Figure~\ref{fig:survey1} highlights a notable improvement in safety perception ($+16. 7\%$), operational confidence ($+23. 3\%$) and overall preference (from $50\%$ to $76.7\%$) when anomaly detection was enabled.

\begin{table}[h!]
\centering
\begin{tabular}{|p{3.5cm}|c|c|c|}
\hline
\textbf{Question} & \textbf{Yes (\%)} & \textbf{Neutral (\%)} & \textbf{No (\%)} \\
\hline
The robot's anomaly detection helped me 
feel safer during operation & 73.3\% & 20\% & 6.7\% \\
\hline
The information provided by the 
robot was clear and useful & 76.7\% & 16.7\% & 6.6\% \\
\hline
The robot's anomaly detection increased
my confidence in its safety & 66.7\% & 16.7\% & 16.7\% \\
\hline
I felt more secure when

anomaly detection was active & 70\% & 26.7\% & 3.3\% \\
\hline
\end{tabular}
\caption{Survey results measuring user safety perception and confidence in anomaly detection}
\label{tab:survey_results}
\end{table}

\subsection{Anomaly Detection Latency Analysis}
The latency from module initialization to LLM anomaly classification display was measured in 88 samples, ranging from 5.99 to 50.69 seconds, with an average of 20 seconds. To improve consistency, BLIP captioning and LLM inference were offloaded to external microservices, reducing the load on the Raspberry Pi 4 (4GB Random Access Memory). Manual activation of the anomaly detection pipeline also reduced delays compared to fixed 25-second intervals. At higher latencies, anomaly classifications were typically delivered after the robot's action, but with optimization, they more often occurred during the action, enhancing safety and responsiveness. Higher latency directly impacts the anomaly detection factor \( \varepsilon \), as delayed detections reduce system safety, perceived responsiveness, and operational reliability. Thus, minimizing latency is critical to maintaining high \( \varepsilon \) scores in safety-critical applications.


\begin{figure}[th!]
    \centering
    \includegraphics[width=1\linewidth]{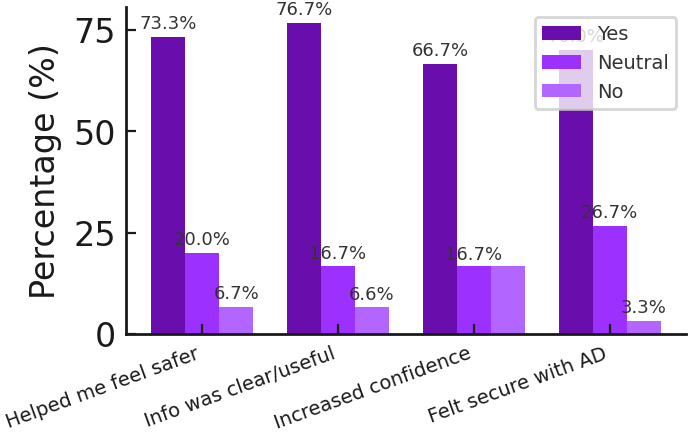}
    \caption{Survey results on user perceptions of anomaly detection, showing that the majority of participants reported increased safety, clarity of information, and confidence when the anomaly detection system was active.}
    \label{fig:survey1}
\end{figure}


\subsection{Response Consistency via Confusion Matrix}
We evaluated the precision of the anomaly detection by comparing the model output with ground truth labels. 
These labels were annotated on the basis of human interpretations of environmental conditions and potential hazards. The confusion matrix thus evaluates how closely the system's anomaly classifications align with human assessments, making it a proxy for measuring detection accuracy.
Table~\ref{tab:performance_metrics} shows the confusion matrix with 196 evaluated images ($TP = 82$, $FN = 15$, $FP = 20$, $TN = 79$), where TP represents True Positives, FP represents False Positives, FN represents False Negatives, and TN represents True Negatives. Performance metrics were computed as follows.

\begin{equation}
\text{Accuracy} = \frac{TP + TN}{TP + FP + FN + TN} = 82.14\%,
\label{eq:accuracy}
\end{equation}

\begin{table}[h!]
\centering
\begin{tabular}{|c|c|c|}
\hline
\textbf{} & \textbf{Predicted Positive} & \textbf{Predicted Negative} \\
\hline
\textbf{Actual Positive} & TP: 82 & FN: 15 \\
\textbf{Actual Negative} & FP: 20 & TN: 79 \\
\hline
\end{tabular}
\caption{Confusion matrix showing performance of the anomaly detection module. True Positive (TP), False Positive (FP), False Negative (FN), True Negative (TN).}
\label{tab:performance_metrics}
\end{table}

\subsection{Anomaly Detection Latency and Prediction Accuracy}

To evaluate real-time responsiveness and anomaly detection accuracy of the system, we extended our experiments with another latency logging module and prediction validation logic using logs collected from 125 trials.

\subsubsection{Latency Distribution}
Latency was measured as the time taken from image capture to anomaly classification delivery. Data from 125 tests were categorized into four latency boxes: 0–8\,s, 8–14\,s, 14–20\,s and 20–26\,s. Most of the responses (84\%) occurred within the range of 0 to 14\,s, indicating strong real-time capabilities. The minimum, maximum, and average latencies were:

\begin{itemize}
    \item \textbf{Minimum Latency:} 2.457 s
    \item \textbf{Maximum Latency:} 25.196 s
    \item \textbf{Average Latency:} 6.017 s
\end{itemize}

\begin{figure}
    \centering
    \includegraphics[width=1\linewidth]{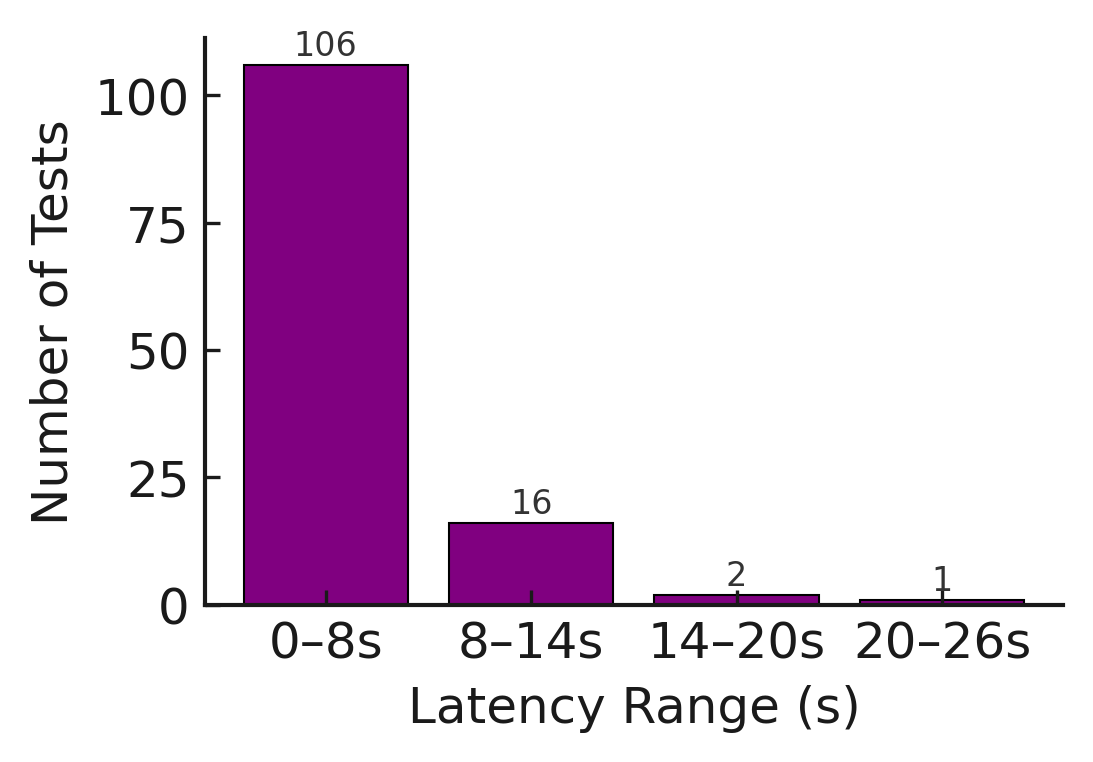}
    \caption{Distribution of Anomaly Detection Latencies across 125 tests.}
    \label{fig:placeholder}
\end{figure}
Compared to earlier evaluation runs (with average latency $\sim$20\,s), this improved latency profile suggests improved performance by offloading LLM inference and BLIP processing to external services.

\subsubsection{Prediction Accuracy of Anomaly Detection}
Of 125 total tests, the system generated 114 correct anomaly detections, with 11 mismatches identified by manual verification, resulting in a detection accuracy of 91.2\%. The detection rate was calculated using the following.

\begin{equation}
\text{Detection Rate} = \frac{U + 0.5N}{T} \times 100
\end{equation}

where $U$ is the number of correct predictions, $N$ is the number of neutral predictions (zero in this case), and $T$ is the total number of trials.

These results demonstrate a strong alignment between the system's predictive behavior and human judgment, reinforcing the reliability of the anomaly detection module in dynamic environments. Although high accuracy indicates robust performance, further refinement may help reduce the occurrence of false positives and improve safety in ambiguous scenarios.

\section{CONCLUSIONS} 
\label{sec:conclusion}
This study demonstrates that the integration of proactive anomaly detection through visual and language models enhances the safety and operational reliability of autonomous mobile robot navigation. The proposed multimodal system improves both navigation performance and safety outcomes in dynamic environments. By extending this framework to comprehensive anomaly detection scenarios, we further showed that robots can not only identify but also proactively respond to environmental threats, enabling real-world safety applications such as hazard detection, conflict resolution, and emergency response. Survey responses and experimental evaluations confirm that proactive anomaly detection fosters operational confidence, safety perception, and system reliability by aligning robot behavior with human safety expectations and reducing operational risks. The high prediction accuracy of the system validates its effectiveness in addressing the critical need for autonomous safety systems. Although latency remains a consideration, results show that optimized anomaly detection delivery contributes to more predictable and safety-oriented robotic actions. The updated system achieved a prediction accuracy of 91.2\% and demonstrated significantly improved latency performance, with 85\% of the responses occurring in 8 seconds. These findings reinforce the system's potential for real-world deployment, where rapid and reliable anomaly detection is essential for safe and autonomous robot operation in complex, dynamic environments.

\section*{Acknowledgments}
The authors thank Nokia Bell Labs and Mike Coss for their invaluable feedback. 
We also thank Prof. Katsuo Kurabayashi and Dr. Rui Li for their insightful comments and guidance. We appreciate the support and encouragement of our colleagues in the Department of Mechanical and Aerospace Engineering at New York University.

\bibliographystyle{unsrt}
\bibliography{references}

\end{document}